\documentclass[10pt,twocolumn,letterpaper]{article}

\usepackage{cvpr}
\usepackage{times}
\usepackage{epsfig}
\usepackage{graphicx}
\usepackage{amsmath}
\usepackage{amssymb}
\usepackage{algorithm}
\usepackage[noend]{algpseudocode}

\usepackage{subcaption}
\DeclareCaptionSubType*[Alph]{table}
\DeclareCaptionLabelFormat{mystyle}{Table~\bothIfFirst{#1}{ ̃}#2}
\usepackage{color}
\usepackage{xcolor}
\usepackage{comment}

\newcommand{\beginsupplement}{%
        \setcounter{table}{0}
        \renewcommand{\thetable}{S\arabic{table}}%
        \setcounter{figure}{0}
        \renewcommand{\thefigure}{S\arabic{figure}}%
     }

\DeclareMathOperator*{\argmin}{argmin}
\DeclareMathOperator*{\argmax}{argmax}
\newcommand\kmeans{\mathop{\mbox{$k$-$\mathit{means}$}}}

\usepackage[pagebackref=true,breaklinks=true,letterpaper=true,colorlinks,bookmarks=false]{hyperref}

 \cvprfinalcopy 

\begin{document}

\title{Cognitively-Inspired Model for Incremental Learning Using a Few Examples}
\author{Ali Ayub, Alan R. Wagner\\
The Pennsylvania State University\\
{\tt\small \{aja5755,alan.r.wagner\}@psu.edu}}

\maketitle
\begin{abstract}
\label{sec:Abstract}
   Incremental learning attempts to develop a classifier which learns continuously from a stream of data segregated into different classes. Deep learning approaches suffer from catastrophic forgetting when learning classes incrementally, while most incremental learning approaches require a large amount of training data per class. We examine the problem of incremental learning using only a few training examples, referred to as Few-Shot Incremental Learning (FSIL). To solve this problem, we propose a novel approach inspired by the concept learning model of the hippocampus and the neocortex that represents each image class as centroids and does not suffer from catastrophic forgetting. We evaluate our approach on three class-incremental learning benchmarks: Caltech-101, CUBS-200-2011 and CIFAR-100 for incremental and few-shot incremental learning and show that our approach achieves state-of-the-art results in terms of classification accuracy over all learned classes.\footnote{Code available at: https://github.com/aliayub7/CBCL}
\end{abstract}
\section{Introduction}
\label{sec:Intro}
Humans can continuously learn new concepts over their lifetime. In contrast, modern machine learning systems often must be trained on batches of data~\cite{Girshick_2015_ICCV,Long_2015_CVPR}. Applied to the task of object recognition, \textit{incremental learning} is an avenue of research that seeks to develop systems that are capable of continually updating the learned model as new data arrives~\cite{Hou_2019_CVPR}. Incremental learning gradually increases an object classifier's breadth by training it to recognize new object classes~\cite{Dhar_2019_CVPR}. 

This paper examines a sub-type of incremental learning known as \textit{class-incremental learning}. Class-incremental learning attempts to first learn a small subset of classes and then incrementally expand that set with new classes. Importantly, class-incremental evaluation of a final model is tested on a single blended dataset, an evaluation known as single-headed evaluation~\cite{Chaudhry_2018_ECCV}. To paraphrase Rebuffi et al.~\cite{Rebuffi_2017_CVPR}, a class-incremental learning algorithm must:  
\begin{enumerate}
    \item Be trainable from a stream of data that includes instances of different classes at different times; 
    \item Offer a competitively accurate multi-class classifier for any classes it has observed thus far; 
    \item Be bounded or only grow slowly with respect to memory and computational requirements as the number of training classes increase.
\end{enumerate}

Creating a high accuracy classifier that incrementally learns, however, is a hard problem. One simple way to create an incremental learner is by tuning the model to the data of the new classes. This approach, however, causes the model to forget the previously learned classes and the overall classification accuracy decreases, a phenomenon known as \textit{catastrophic forgetting}~\cite{french19,kirkpatrick17}. To overcome this problem, most existing class-incremental learning methods avoid it altogether by storing a portion of the training data from the earlier learned classes and retrain the model (often a neural network) on a mixture of the stored data and new data containing new classes~\cite{Rebuffi_2017_CVPR,Castro_2018_ECCV,Wu_2019_CVPR}. These approaches are, however, neither scalable nor biologically inspired i.e. when humans learn new visual objects they do not forget the visual objects they have previously learned, nor must humans relearn these previously known objects. Furthermore, current methods for incremental learning require a large amount of training data and are thus not suitable for training from a small set of examples. 

We seek to develop a practical incremental learning system that would allow human users to incrementally teach a robot different classes of objects. In order to be practical for human users, an incremental learner should only require a few instances of labeled data per class. Hence, in this paper we explore the Few-Shot Incremental Learning (FSIL) problem, in which an agent/robot is required to learn new classes continually but with only a small set of examples per class. 

With respect to class-incremental learning and FSIL, this paper contributes a novel cognitively-inspired method termed Centroid-Based Concept Learning (CBCL). CBCL is inspired by the concept learning model of the hippocampus and the neocortex~\cite{Mack18,renoult15,moscovitch16}. CBCL treats each image as an episode and extracts its high-level features. CBCL uses a fixed data representation (ResNet~\cite{He_2016_CVPR} pre-trained on ImageNet~\cite{Russakovsky15}) for feature extraction. After feature extraction, CBCL generates a set of concepts in the form of centroids for each class using a cognitively-inspired clustering approach (denoted as \textit{Agg-Var} clustering) proposed in \cite{Ayub_scenes20}. After generating the centroids, to predict the label of a test image, the distance of the feature vector of the test image to the $n$ closest centroids is used. Since CBCL stores the centroids for each class independently of the other classes, the decrease in overall classification accuracy is not catastrophic when new classes are learned. CBCL is tested on three incremental learning benchmarks (Caltech-101~\cite{fei-fei06}, CUBS-200-2011~\cite{wah11}, CIFAR-100~\cite{Krizhevsky09}) and it outperforms the state-of-the-art methods by a sizable margin. Evaluations for FSIL show that CBCL outperforms some class-incremental learning methods, even when CBCL uses only 5 or 10 training examples per class and other methods use the complete training set per class (500 images per class for CIFAR-100). For FSIL, CBCL even beats a few-shot learning baseline that learns from the training data of all classes (batch learning) on the three benchmark datasets. The main contributions of this paper are:

\begin{enumerate}
    \item A cognitively-inspired class-incremental learning approach is proposed that outperforms the state-of-the-art methods on the three benchmark datasets listed above. 
    \item A novel centroid reduction method is proposed that bounds the memory footprint without a significant loss in classification accuracy.
    \item A challenging incremental learning problem is examined (FSIL) and experimental evaluations show that our approach results in state-of-the-art accuracy when applied to this problem. 
\end{enumerate}

\section{Related Work}
\label{sec:related_work}
The related work is divided into two categories: traditional approaches that use a fixed data representation and class-incremental approaches that use deep learning. 

\subsection{Traditional Methods}

Early incremental learning approaches used SVMs~\cite{Cortes95}. For example, Ruping~\cite{ruping01} creates an incremental learner by storing support vectors from previously learned classes and using a mix of old and new support vectors to classify new data. Most of the earliest approaches did not fulfill the criteria for class-incremental learning and many required old class data to be available when learning new classes:~\cite{kuzborskij13,Pentina_2015_CVPR,polikar01,muhlbaier09}.

Another set of early approaches use a fixed data representation with a \textit{Nearest Class Mean} (NCM) classifier for incremental learning~\cite{Mensink13,Mensink12,Ristin_2014_CVPR}. NCM classifier computes a single centroid for each class as the mean of all the feature vectors of the images in the training set for each class. To predict the label for a test image, NCM assigns it the class label of the closest centroid. NCM avoids catastrophic forgetting by using centroids. Each class centroid is computed using only the training data of that class, hence even if the classes are learned in an incremental fashion the centroids for previous classes are not affected when new classes are learned. These early approaches, however, use SIFT features ~\cite{Lowe04}, hence their classification accuracy is not comparable to the current deep learning approaches, as shown in~\cite{Rebuffi_2017_CVPR}.  

\subsection{Deep Learning Methods}
Deep learning methods have produced excellent results on many vision tasks  because of their ability to jointly learn task-specific features and classifiers~\cite{Bengio13,Girshick_2015_ICCV,Long_2015_CVPR,Simonyan14}. However, deep learning approaches suffer from catastrophic forgetting on incremental learning tasks. Essentially, classification accuracy rapidly decreases when learning new classes~\cite{ans04,french19,Goodfellow13,kirkpatrick17,lee17,mccloskey89}. Various approaches have been proposed recently to deal with catastrophic forgetting for task-incremental and class-incremental learning~\cite{Aljundi_2018_ECCV,Rebuffi_2017_CVPR}. 

For task-incremental learning, a model is trained incrementally on different datasets and during evaluation it is tested on the different datasets separately~\cite{Chaudhry_2018_ECCV}. Task-incremental learning utilizes multi-headed evaluation which is characterized by predicting the class when the task is known, which has been shown to be a much easier problem in~\cite{Chaudhry_2018_ECCV} than the class-incremental learning considered in this paper.

\subsubsection{Class-Incremental Learning Methods}
Most of the recent class-incremental learning methods rely on storing a fraction of old class data when learning a new class~\cite{Rebuffi_2017_CVPR,Hou_2019_CVPR,Castro_2018_ECCV,Wu_2019_CVPR,Chaudhry_2018_ECCV}. iCaRL~\cite{Rebuffi_2017_CVPR} combines knowledge distillation~\cite{Hinton15} and NCM for class-incremental learning. Knowledge distillation uses a distillation loss term that forces the labels of the training data of previously learned classes to remain the same when learning new classes. iCaRL uses the old class data while learning a representation for new classes and uses the NCM classifier for classification of the old and new classes. EEIL~\cite{Castro_2018_ECCV} improves iCaRL with an end-to-end learning approach. Hou et al.~\cite{Hou_2019_CVPR} uses cosine normalization, less-forget constraint and inter-class separation for reducing the data imbalance between old and new classes. The main issue with these approaches is the need to store old class data which is not practical when the memory budget is limited. To the best of our knowledge, there are only two approaches that do not use old class data and use a fixed memory budget: LWF-MC~\cite{Rebuffi_2017_CVPR} and LWM~\cite{Dhar_2019_CVPR}. LWF-MC is simply the implementation of LWF~\cite{Li18} for class-incremental learning. LWM uses attention distillation loss and a teacher model trained on old class data for better performance than LWF-MC. Although both of these approaches meet the conditions for class-incremental learning proposed in \cite{Rebuffi_2017_CVPR}, their performance is inferior to approaches that store old class data~\cite{Rebuffi_2017_CVPR,Castro_2018_ECCV,Wu_2019_CVPR}. 

An alternative set of approaches increase the number of layers in the network for learning new classes~\cite{Rusu16,Terekhov15}. Another novel approach is presented in~\cite{Xiao14} which grows a tree structure to incorporate new classes incrementally. These approaches also have the drawback of rapid increase in memory usage as new classes are added. 

Some researchers have also focused on using a deep network pre-trained on ImageNet as a fixed feature extractor for incremental learning. Belouadah et al.~\cite{Belouadah_2018_ECCV_Workshops} uses a pre-trained network for feature extraction and then trains shallow networks for classification while incrementally learning classes. They also store a portion of old class data. The main issue with their approach is that they test their approach on the ImageNet dataset using the feature extractor that has already been trained on ImageNet which skews their results. FearNet~\cite{kemker18} uses a ResNet-50 pre-trained on ImageNet for feature extraction and uses a brain-inspired dual memory system which requires storage of the feature vectors and co-variance matrices for the old class images. The feature vectors and co-variance matrices are further used for generating augmented data during learning. Our approach does not store any base class data or use any data augmentation, although it uses a ResNet pre-trained on ImageNet for feature extraction but we do not test our approach on ImageNet.


\begin{figure*}[t]
\centering
\includegraphics[scale=0.365]{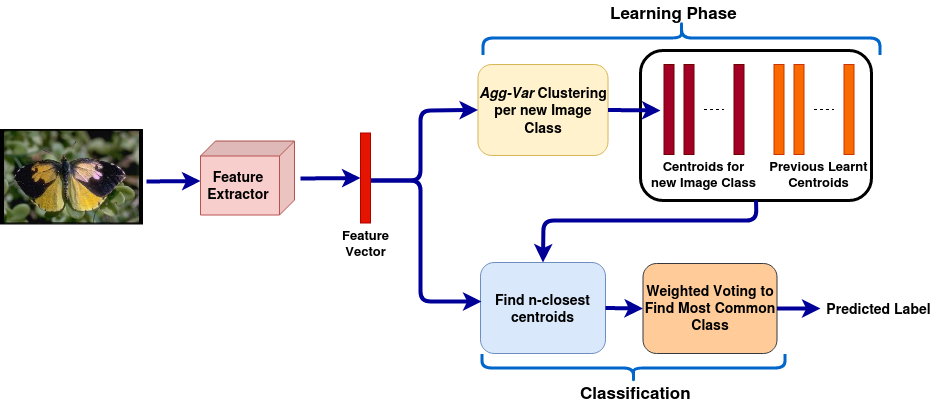}
\caption{\small For each new image class in a dataset, the feature extractor generates the CNN features of all the training images in the image class and generates a set of centroids using \textit{Agg-Var} clustering algorithm, concatenates them with the centroids of previously learned classes and uses the complete set of centroids for classifying unlabeled test images}
\label{fig:framework}
\end{figure*}
\section{Methodology}
\label{sec:Method}

Following the notation from~\cite{Rebuffi_2017_CVPR}, CBCL learns from a class-incremental data stream of sample sets $X^1, X^2, ..., X^N$ in which all samples from the set $X^y=\{x_1^y, ..., x_{N_y}^y\}$ are from the class $y\in{N}$ with $N_y$ samples. 

The subsections below, first explain our method for class-incremental learning. Next, we explore how the memory footprint can be managed by restricting the total number of centroids. Finally, we demonstrate the use of our approach to incrementally learn using only a few examples per class.  

\subsection{\textbf{\textit{Agg-Var}} Clustering}
The complete architecture of our approach is depicted in Figure \ref{fig:framework}. Once the data for a new class becomes available, the first step in CBCL is the generation of feature vectors from the images of the new class using a fixed feature extractor. The proposed architecture can work with any type of image feature extractor or even for non-image datasets with appropriate feature extractors. In this paper, for the task of object-centric image classification, we use CNNs (ResNet~\cite{He_2016_CVPR}) pre-trained on ImageNet~\cite{Russakovsky15} as feature extractors. 

In the learning phase, for each new image class $1 \leq y \leq N$, \textit{Agg-Var} clustering \cite{Ayub_scenes20} is applied on the feature vectors of all the training images in the class $\{x_1^y, x_2^y, ..., x_{N_y}^y\}$. In the hippocampal concept learning model~\cite{Mack18,renoult15,moscovitch16}, after the feature extraction step, the hippocampus calculates a term called the \textit{memory-based prediction error}. This value represents the difference from the incoming episode to all of the previously experienced concepts. This step is replicated in \textit{Agg-Var} clustering by finding the Euclidean distance between the incoming image to each centroid for a class. Initially there are no centroids for a new class $y$. Hence, this step begins by creating a centroid from the first image of class $y$. Next, for each image in training set of the class, feature vector $x_i^y$ (for the $i$th image) is generated and compared using the Euclidean distance to all the centroids for the class $y$. If the distance of $x_i^y$ to the closest centroid is below a pre-defined distance threshold $D$, the closest centroid is updated by calculating a weighted mean of the centroid and the feature vector $x_i^y$:
\begin{equation}
C_{new} = \frac{w_C \times C_{Old} + x_i^y}{w_C+1}
\end{equation}
\noindent where, $C_{new}$ is the updated centroid, $C_{old}$ is the centroid before the update, $w_C$ is the number of data points (images) already represented by the centroid. This step of \textit{Agg-Var} clustering is meant to capture \textit{memory integration} in the concept learning process of the hippocampus. Memory integration occurs when the \textit{memory-based prediction error} of an episode to a previous concept is small. If, on the other hand, the \textit{memory-based prediction error} of an episode to a previous concept is large, according to the concept learning process of the hippocampus, \textit{pattern separation} occurs resulting in the creation of a new distinct concept based on the incoming episode. \textit{Agg-Var} clustering captures this aspect of the process as: if the distance between the $i$th image and the closest centroid is higher than the distance threshold $D$, a new centroid is created for class $y$ and equated to the feature vector $x_i^y$ of the $i$th image. The result of this process is a collection containing a set of centroids for the class $y$, $C^y = \{c_1^y, ..., c_{N^*_y}^y\}$, where $N^*_y$ is the number of centroids for class $y$. This process is applied to the sample set $X^y$ of each class incrementally once they become available to get a collection of centroids $C  = C^1, C^2, ..., C^N$ for all $N$ classes in a dataset. It should be noted that using the same distance threshold for different classes can yield different number of centroids per class depending on the similarity among the images (intra-class variance) in each class. Hence, we only need to tune a single parameter ($D$) to get the optimal number of centroids that yield best validation accuracy in each class. Note that our approach calculates the centroids for each class separately. Thus, the performance of our approach is not strongly impacted when the classes are presented incrementally. 

\subsection{\textbf{Weighted-Voting Scheme for Classification}}
To predict the label $y^*$ of a test image we use the feature extractor to generate a feature vector $x$. Next, Euclidean distance is calculated between $x$ and the centroids of all the classes observed so far. Based on the calculated distances, we select $n$ closest centroids to the unlabeled image. The contribution of each of the $n$ closest centroids to the determination of the test image's class is a conditional summation:

\begin{equation}
    Pred(y) = \sum_{j=1}^{n} \frac{1}{dist(x,c_j)}[y_j=y]
\end{equation}

\noindent where $Pred(y)$ is the prediction weight of class $y$, $y_j$ is category label of $j$th closest centroid $c_j$ and $dist(x,c_j)$ is the euclidean distance between $c_j$ and the feature vector $x$ of the test image. The prediction weights for all the image classes observed so far are first initialized to zero. Then, for the $n$ closest centroids the prediction weights are updated, using equation (2), for the classes that each of the $n$ centroids belong to. The prediction weight for each class is further multiplied by the inverse of the total number of images in the training set of the class to manage class imbalance. Since classes with more training data most likely have more centroids than other classes, prediction weight can become biased towards such classes. The proposed weighting scheme avoid bias towards such classes during prediction:
\begin{equation}
    Pred(y)^{'} = \frac{1}{N_y}Pred(y)
\end{equation}
where $Pred(y)^{'}$ is the prediction weight of class $y$ after multiplication with the inverse of total number of training images of the class $N_y$ with the previous prediction weight $Pred(y)$ of the class. The test image is assigned the class label with the highest prediction weight $Pred(y)^{'}$. 

\subsection{Centroid Reduction}
The memory footprint is an important consideration for an incremental learning algorithm~\cite{Rebuffi_2017_CVPR}. Real system implementations have limited memory available. We therefore propose a novel method that restricts the number of centroids while attempting to maintain classification accuracy. 

If we assume that a system can store a maximum of $K$ centroids and that currently the system has stored $K_t$ centroids for $t$ classes. For the next batch of classes the system needs to store $K_{new}$ more centroids but the total number of centroids $K_t + K_{new} > K$. Hence, the system needs to reduce the total stored centroids to $K_r = K_t+K_{new}-K$ centroids. Rather than reducing the number of centroids for each class equally, CBCL reduces the centroids for each class based upon the previous number of centroids in the class. The reduction in the number of centroids $N_y^*$ for each class $y$ is calculated as (whole number):
\begin{equation}
    N^*_{y}(new) = N_y^*(1-\frac{K_r}{K_t})
\end{equation}

\noindent where $N_y^*(new)$ is the number of centroids for class $y$ after reduction. 
Rather than simply removing the extra centroids from each class, we cluster the closest centroids in each class to get new centroids, keeping as much information as possible about the previous classes. This process is accomplished by applying k-means clustering~\cite{jain99} on the centroid set $C^y$ of each class $y$ to cluster them into a total of $N_y^*(new)$ centroids. Results on benchmark datasets show the effectiveness of our centriod reduction approach (Section \ref{sec:Experiments}).  

\subsection{Few-Shot Incremental Learning (FSIL)}
For a traditional few-shot learning problem, an algorithm is evaluated on n-shots, k-way tasks. Hence, a model is given a total of $n$ examples per class for $k$ classes for training. After the training phase, the model is evaluated on a small number of test samples (usually 15 test samples for 1-shot, 5-shot and 10-shot learning) for each of the $k$ classes. Some few-shot learning approaches have been proposed in which the model is tested on the new $k$ classes and the base classes \cite{Gidaris_2018_CVPR,Qi_2018_CVPR,ren19}. However, these approaches are not suitable for learning classes incrementally for a large number of increments using only a few samples per class. The few-shot incremental learning setting proposed here deals with this problem. 

For an n-shot incremental learning setting, we propose to train a model on $n$ examples per class for $k$ classes in an increment. The training data for the $l$ previously learned classes is not available to the model during the current increment. After training, the model is tested on the complete test set for all the classes learned so far ($k+l$).  Although this problem becomes more difficult with each increment, we show that our approach performs well even for 5-shot and 10-shot incremental learning cases (Section \ref{sec:Experiments}) because even a limited number of instances per class generate centroids covering most of the class's concept. FSIL is potentially important for applications where labeled data is difficult to obtain, perhaps such as a human incrementally teaching a robot. In such cases, the human is unlikely to be willing to provide more than few examples of a class.

\section{Experiments}
\label{sec:Experiments}
We evaluate CBCL on three standard class-incremental learning datasets: Caltech-101 \cite{fei-fei06}, CUBS-200-2011 \cite{wah11} and CIFAR-100 \cite{Krizhevsky09}. First, we present the datasets and the implementation details. CBCL is then compared to state-of-the-art methods for class-incremental learning and evaluated on 5-shot and 10-shot incremental learning. Finally, we perform an ablation study to analyze the contribution of each component of our approach. 

\begin{table}[t]
\centering
\small
\begin{tabular}{ p{2.2cm}p{1.38cm}p{1.5cm}p{1.5cm} }
     \hline
    \textbf{Dataset} & \textbf{CIFAR-100} & \textbf{CUBS-200-2011} & \textbf{Caltech-101} \\
     \hline
    \# classes & 100 & 100 & 100\\
     \hline
    \# training images &  500 & 80\% of data & 80\% of data\\
    \# testing images &  100 & 20\% of data & 20\% of data\\
    \# classes/batch & 2, 5, 10, 20 & 10 & 10\\
 \hline
 \end{tabular}
 \caption{Statistical details of the datasets in our experiments, same as in~\cite{Dhar_2019_CVPR} for a fair comparison. Number of training and test images reported are for each class in the dataset.}
 \label{tab:Datasets}
 \end{table}

\subsection{Datasets}
CBCL was evaluated on the three datasets used in \cite{Dhar_2019_CVPR}. LWM~\cite{Dhar_2019_CVPR} was also tested on iLSVRC-small(ImageNet) dataset but since our feature extractor is pre-trained on ImageNet, comparing on this dataset would not be a fair comparison. Caltech-101 contains 8,677 images of 101 object categories with 40 to 800 images per category. CUBS-200-2011 contains 11,788 images of 200 categories of birds. CIFAR-100 consists of 60,000 $32\times32$ images belonging to 100 object classes. There are 500 training images and 100 test images for each class. The number of classes, train/test split size and number of classes per batch used for training are described in Table~\ref{tab:Datasets}. The classes that compose a batch were randomly selected. For the 5-shot and 10-shot incremental learning experiments, only the training images per class were changed to 5 and 10, respectively in Table~\ref{tab:Datasets} keeping the other statistics the same. 

Similar to~\cite{Dhar_2019_CVPR,Rebuffi_2017_CVPR}, top-1 accuracy was used for evaluation. We also report the \textit{average incremental accuracy}, which is the average of the classification accuracies achieved in all the increments~\cite{Rebuffi_2017_CVPR}. 

\begin{figure*}
\centering
\includegraphics[scale=0.38]{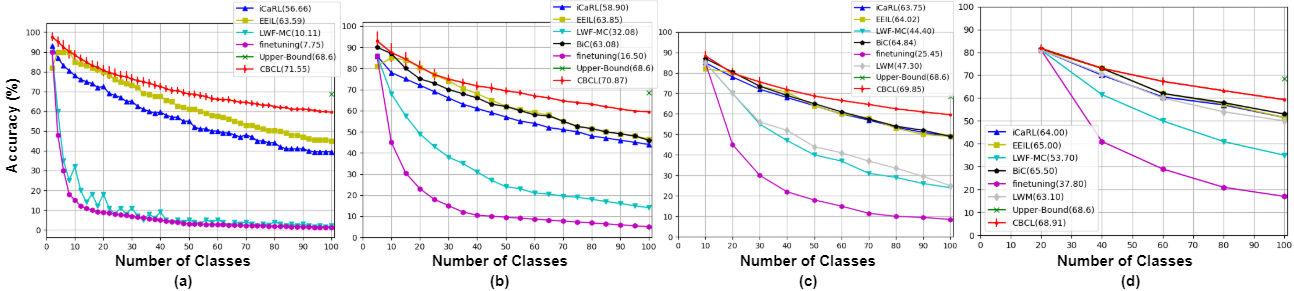}
\caption{\small Average and standard deviation of classification accuracies (\%) on CIFAR-100 dataset with (a) 2, (b) 5, (c) 10, (d) 20 classes per increment with 10 executions. Average incremental accuracies are shown in parenthesis. (For other methods, results are reported from the respective papers and different papers reported results on different increment settings. Best viewed in color)}
\label{fig:cifar}
\end{figure*}


Because CBCL's learning time is much shorter than the time required to train a neural network, we are able to run all our experiments 10 times randomizing the order of the classes. We report the average classification accuracy and standard deviation over these ten runs.

\subsection{Implementation Details}
The Keras deep learning framework~\cite{chollet2015} was used to implement all of the neural network models. For Caltech-101 and CUBS-200-2011 datasets, the ResNet-18~\cite{He_2016_CVPR} model pre-trained on the ImageNet~\cite{Russakovsky15} dataset was used and for CIFAR-100 the ResNet-34~\cite{He_2016_CVPR} model pre-trained on ImageNet was used for feature extraction. These model architectures are consistent with~\cite{Dhar_2019_CVPR} for a fair comparison. For the experiment with the CIFAR-100 dataset the model was allowed to store up to $K=7500$ centroids requiring 3.87 MB versus 84 MB for an extra ResNet-34 teacher model as in~\cite{Dhar_2019_CVPR}. Furthermore, compared to methods that store only 2000 images for previous classes~\cite{Rebuffi_2017_CVPR,Castro_2018_ECCV,Belouadah_2018_ECCV_Workshops}, 7500 centroids for our approach require less memory (3.87 MB) than 2000 complete images (17.6 MB). For Caltech-101 $K=1100$ centroids were stored (0.5676 MB versus 45 MB for a ResNet-18 teacher model as in~\cite{Dhar_2019_CVPR}) and for CUBS-200-2011 $K=800$ centroids were stored (0.4128 MB). 

\begin{figure}[t]
\centering
\includegraphics[scale=0.3]{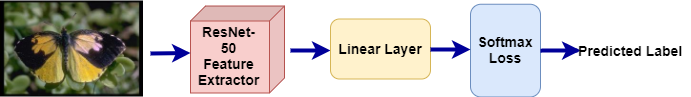}
\caption{\small Few-shot Learning Baseline (FLB) architecture}
\label{fig:baseline}
\end{figure}

As mentioned in Section \ref{sec:Intro}, none of the prior incremental learning techniques are suitable for FSIL because they require a large amount of training data per class. Hence, we compare CBCL against a few-shot learning baseline (FLB). FLB uses the features from the pre-trained ResNet neural network which are passed on to a linear layer which is trained with softmax loss (Figure \ref{fig:baseline}). This procedure follows prior work on few-shot learning research \cite{Chen19}, which indicates that FLB is better than many other few-shot learning techniques that use a deeper backbone, such as ResNet-18 or ResNet-34. Since FLB is not suitable for few-shot incremental learning, we train the final linear layer of FLB with softmax loss using the complete training set of all the new and old class data in each increment. In other words, FLB does not learn incrementally. FLB was trained for 25 epochs in each increment using a fixed learning rate of 0.001 and cross-entropy loss with minibatches of size 8 optimized using stochastic gradient descent.

For CBCL, for each batch of new classes, the hyper-parameters $D$ (distance threshold) and $n$ (number of closest centroids used for classification) are tuned using cross-validation. We only use the previously learned centroids and the training data of the new classes for hyper-parameter tuning.

\subsection{Results on CIFAR-100 Dataset}
\label{sec:Cifar}
On CIFAR-100 dataset, our method is compared to seven different methods: \textit{finetuning} (FT), LWM~\cite{Dhar_2019_CVPR}, LWF-MC~\cite{Rebuffi_2017_CVPR}, iCaRL~\cite{Rebuffi_2017_CVPR}, EEIL~\cite{Castro_2018_ECCV}, BiC~\cite{Wu_2019_CVPR} and FearNet~\cite{kemker18}\footnote{Comparison with FearNet is in Supplementary File}. FT simply uses the network trained on previous classes and adapts it to the new incoming classes. LWM extends LWF~\cite{Li18} and uses attention distillation loss for class-incremental learning. 
LWF-MC uses distillation loss during the training phase. iCaRL also uses the distillation loss for representation learning but stores exemplars of previous classes and uses the NCM classifier for classification. EEIL improves iCaRL by offering an end-to-end learning approach which also uses the distillation loss and keeps exemplars from the old classes. BiC also uses the exemplars from the old classes and adds a bias correction layer after the fully connected layer of the ResNet to correct for the bias towards the new classes. We also compare the classification accuracy after learning all the classes to an upper bound (68.6\%) consisting of ResNet34 trained on the entire CIFAR-100 dataset in one batch. 

\begin{table}[t]
\centering
\small
\begin{tabular}{ |p{1.2cm}|p{0.9cm}|p{0.65cm}|p{0.65cm}|p{0.65cm}|p{0.65cm}| }
    \hline
    \multicolumn{2}{|c}{} & \multicolumn{4}{|c|}{\textbf{Classes per increment}}\\
     \hline
    \textbf{Methods} &\textbf{k-Shot} & \textbf{2} & \textbf{5} & \textbf{10} &\textbf{20} \\
    \hline
    FLB & 5 & 41.1 & 39.9 & 41.3 & 44.4 \\
    \hline
    \textbf{CBCL} & 5 & \textbf{56.9} & \textbf{55.6} & \textbf{54.7} & \textbf{53.8} \\
    \hline
    FLB & 10 & 53.5 & 52.4 & 55.1 & 55.6 \\
    \hline
    \textbf{CBCL} & 10 & \textbf{61.9} & \textbf{61.4} & \textbf{61.3} & \textbf{60.7} \\
 \hline
 \end{tabular}
 \caption{Comparison of CBCL with FLB on 5-shot and 10-shot incremental learning settings in terms of average incremental accuracy (\%) on CIFAR-100 dataset with 2, 5, 10 and 20 classes per increment.}
 \label{tab:average_increment}
 \end{table}

Figure~\ref{fig:cifar} compares CBCL to first six out of the seven methods mentioned above with 2, 5, 10 and 20 classes per increment. Even though a fair comparison of CBCL is only possible with FT, LWF-MC and LWM, since they are the only approaches that do not require storing the exemplars of the old classes, it outperforms all six methods on all increment settings. The difference in classification accuracy between CBCL and these other methods increases as the number of classes learned increases. Moreover, for smaller increments the difference in accuracy is larger. Unlike other methods, CBCL's performance remains the same regardless of the number of classes in each increment (final accuracy after 100 classes for all increments is \textbf{60\%}).

Table \ref{tab:average_increment} compares CBCL with FLB for 5-shot and 10-shot incremental learning in terms of average incremental accuracy. CBCL beats FLB on both 5-shot and 10-shot incremental learning for all four incremental settings with significant margins. It should be noted that FLB uses the training set of all the old and new classes in each increment while CBCL uses the training set of new classes only. Further, the difference in accuracy between CBCL and FLB is higher when using 5 examples per class. Also, the difference is higher when using smaller number of classes per increment. This may suggest that CBCL is best suited for incremental learning situations when data is scarce.  

Comparing the average incremental accuracies of the other six methods (Figure \ref{fig:cifar}), which use the complete training set per class, even with only 5 or 10 training examples per class CBCL outperforms other methods that do not store class data (FT, LWF-MC, LWM) and is only slightly inferior to methods that do store old class data (2, 5, 10 classes per increment). For 20 classes per increment, CBCL is slightly inferior to the other methods when using 5 and 10 examples per class. For further comparison, when ResNet-34 was trained on 5-shot and 10-shot settings in a single batch it yielded only 8.22\% and 12.15\% accuracies, respectively. These results clearly show that CBCL offers excellent performance on few-shot incremental learning for object classification. 


\begin{table}[t]
\centering
\small
\begin{tabular}{ |p{1.21cm}|p{0.8cm}|p{0.8cm}|p{2.2cm}| }
     \hline
    \textbf{\# Classes} & \textbf{FT} & \textbf{LWM} & \textbf{CBCL (Ours)} \\
     \hline
    10 (base) & $97.78$ & $97.78$ & $\boldsymbol{97.75\pm1.23}$\\
     \hline
    20 & $59.55$ & $75.34$ & $\boldsymbol{95.42\pm2.00}$\\
    \hline
    30 & $52.65$ & $71.78$ & $\boldsymbol{92.58\pm2.08}$\\
    \hline
    40 & $44.51$ & $67.49$ & $\boldsymbol{91.40\pm1.77}$\\
    \hline
    50 & $35.52$ & $59.79$ & $\boldsymbol{90.23\pm1.45}$\\
    \hline
    60 & $31.18$ & $56.62$ & $\boldsymbol{89.34\pm1.57}$\\
    \hline
    70 & $32.99$ & $54.62$ & $\boldsymbol{88.37\pm1.57}$\\
    \hline
    80 & $27.45$ & $48.71$ & $\boldsymbol{87.57\pm1.07}$\\
    \hline
    90 & $28.55$ & $46.21$ & $\boldsymbol{86.99\pm0.95}$\\
    \hline
    100 & $28.26$ & $48.42$ & $\boldsymbol{86.44\pm0.70}$\\
 \hline
 \end{tabular}
 \caption{Comparison with FT and LWM \cite{Dhar_2019_CVPR} on Caltech-101 dataset in terms of classification accuracy (\%) with 10 classes per increment. Average and standard deviation of classification accuracies per increment are reported}
 \label{tab:Caltech}
 \end{table}

\subsection{Results on Caltech-101 Dataset}
\label{sec:Caltech-101}
For the Caltech-101 dataset CBCL was compared to \textit{finetuning} (FT) and LWM \cite{Dhar_2019_CVPR} with learning increments of 10 classes per batch (Table~\ref{tab:Caltech}). FT and LWM were introduced in Subsection~\ref{sec:Cifar}. CBCL outperforms FT and LWM by a significant margin. The difference between CBCL and LWM and FT continues to increase as more classes are learned. FT performs the worst with a classification accuracy after 100 classes that is about one fourth of the base accuracy (decreases by 69.52\%). LWM is an improvement compared to FT. Nevertheless, the accuracy after 100 classes is almost the half of the base accuracy (decreases by 49.36\%). For CBCL the decrease in accuracy is only \textbf{11.31\%} after incrementally learning 100 classes. The average incremental accuracies for FT, LWM, CBCL, CBCL on 5-shot incremental learning and CBCL on 10-shot incremental learning are 43.84\%, 62.67\%, \textbf{90.61\%}, \textbf{87.70\%} and \textbf{89.92\%}, respectively. Hence, CBCL improves accuracy over the current best method (LWM) by a margin of \textbf{27.94\%} in terms of average incremental accuracy when the complete training set is used. Even for the 5-shot and 10-shot incremental learning, CBCL outperforms LWM by margins of \textbf{25.03\%} and \textbf{27.25\%}, respectively. 

We also compare CBCL for 5-shot and 10-shot incremental learning against FLB trained on all the classes data in each increment (batch learning). FLB achieves 72.48\% and 83.81\% average incremental accuracies for 5-shot and 10-shot incremental learning settings, respectively, which are significantly inferior (\textbf{15.22\%} and \textbf{6.11\%}) to CBCL. These results are in accordance with CIFAR-100 results. 

\begin{table}[t]
\centering
\small
\begin{tabular}{ |p{1.21cm}|p{0.8cm}|p{0.8cm}|p{2cm}| }
     \hline
    \textbf{\# Classes} & \textbf{FT} & \textbf{LWM} & \textbf{CBCL (Ours)} \\
     \hline
    10 (base) & $99.17$ & $99.17$ & ${92.83\pm1.78}$\\
     \hline
    20 & $57.92$ & $78.75$ & $\boldsymbol{84.22\pm1.62}$\\
    \hline
    30 & $41.11$ & $70.83$ & $\boldsymbol{74.55\pm3.06}$\\
    \hline
    40 & $35.42$ & $58.54$ & $\boldsymbol{70.73\pm3.26}$\\
    \hline
    50 & $32.33$ & $53.67$ & $\boldsymbol{66.06\pm3.32}$\\
    \hline
    60 & $29.03$ & $47.92$ & $\boldsymbol{62.26\pm2.79}$\\
    \hline
    70 & $22.14$ & $43.79$ & $\boldsymbol{59.58\pm2.54}$\\
    \hline
    80 & $22.27$ & $43.83$ & $\boldsymbol{57.44\pm2.58}$\\
    \hline
    90 & $20.52$ & $39.85$ & $\boldsymbol{56.37\pm2.28}$\\
    \hline
    100 & $17.4$ & $34.52$ & $\boldsymbol{54.75\pm2.66}$\\
 \hline
 \end{tabular}
 \caption{Comparison with FT and LWM \cite{Dhar_2019_CVPR} on CUBS-200-2011 dataset in terms of classification accuracy (\%) with 10 classes per increment. Average and standard deviation of classification accuracies per increment are reported}
 \label{tab:Cubs}
 \end{table}

\subsection{Results on CUBS-200-2011 Dataset}
For the CUBS-200-2011 dataset we again compare our approach to FT and LWM with learning increments of 10 classes per batch (Table~\ref{tab:Cubs}). The classification accuracy of CBCL is greater than FT and LWM after the 10 classes (base). As the learning increments increase the performance margin also increases. The accuracy of FT decreases by 81.77\% after 10 increments and LWM's accuracy decreases by 64.65\%. The decrease in classification accuracy of CBCL after 10 increments is \textbf{38.0\%} lower than both of these approaches. The average incremental accuracies for FT, LWM, CBCL, CBCL for 5-Shot incremental learning and CBCL for 10-shot incremental learning are 37.7\%, 57.0\%, \textbf{67.8\%}, 56.2\% and \textbf{63.8\%} respectively. CBCL is an improvement over LWM by a \textbf{10.7\%} margin in terms of average incremental accuracy. Furthermore, even for 10-shot incremental learning setting CBCL improves over LWM and it is slightly below LWM for 5-shot incremental learning.

Similar to CIFAR-100 and Caltech-101, we compare CBCL on 5-shot and 10-shot incremental learning against FLB trained on all the classes data in each increment (batch learning). FLB achieves 37.48\% and 55.00\% average incremental accuracies for 5-shot and 10-shot incremental learning settings, respectively, which are inferior (\textbf{18.68\%} and \textbf{8.8\%}) to CBCL. These results are in accordance with CIFAR-100 and Caltech-101 FSIL results. 

\subsection{Ablation Study}
We performed an ablation study to examine the contribution of each component in our approach to the overall system's accuracy. This set of experiments was performed on CIFAR-100 dataset with increments of 10 classes and memory budget of $K=7500$ centroids using all the training data per class. We report average incremental accuracy for these experiments.

This ablation study investigates the effect of the following components: feature extractor, clustering approach, number of centroids used for classification, and the impact of centroid reduction. Hybrid versions of CBCL are created to ablate each of these different components. 
Hybrid-1 termed \textit{VGG-16} uses a VGG-16 pre-trained on ImageNet as a feature extractor. Hybrid-2 termed \textit{Trad-Agg} uses traditional agglomerative clustering and hybrid-3 termed \textit{K-means} uses k-means clustering to generate centroids for all the image classes. Hybrid-4 termed \textit{Single-Centroid-Pred} uses only a single closest centroid for classification (same as NCM classifier). Hybrid-5 termed  \textit{Remove-Centroids} simply removes the extra centroids when the memory limit is reached rather than using the proposed centroid reduction technique. Lastly, hybrid-6 termed \textit{NCM} uses an NCM classifier with the ImageNet pre-trained feature extractor. Except for the changed component, all the other components in the hybrid approaches are the same as CBCL. 

Table~\ref{tab:Ablation} shows the results for the ablation study. All of the hybrid methods are less accurate than the complete CBCL algorithm. 
\textit{VGG-16} hybrid achieve slightly lower accuracy than CBCL with ResNet-34, depicting the robustness of our method against the choice of the feature extractor. \textit{Trad-Agg} and \textit{K-means} achieve similar average incremental accuracy but is significantly inferior when compared to CBCL. This difference in accuracy reflects the effectiveness of the \textit{Agg-Var} clustering algorithm for object-centric image classification. \textit{Single-Centroid-Pred} achieves slightly lower accuracy than CBCL, illustrating that the accuracy gain resulting from using multiple centroids for classification is about 1.15\%. Finally, the \textit{Remove-Centroid} hybrid's accuracy is the closest to CBCL's. This small difference may reflect the fact that the memory budget is large enough such that the algorithm does not need to reduce centroids until the last increment. Hence, only in the last increment is there a slight change, which does not effect the average incremental accuracy for all 10 increments by a significant margin. The effectiveness of our centroid reduction technique is more apparent when using smaller memory budgets. For example, for K=3000 centroids limit the average incremental accuracies for CBCL and \textit{Remove-Centroids} are \textbf{67.5\%} and 64.0\%, respectively, depicting the effectiveness of our proposed centroid reduction technique. Lastly, for NCM, we again see a drastic decrease in accuracy because this hybrid uses a single centroid to represent each class. This ablation study clearly indicates that the most important component of CBCL is the cognitively-inspired \textit{Agg-Var} clustering approach, based upon drastic decrease in performance for \textit{Trad-Agg}, \textit{K-means} and \textit{NCM} hybrids. Note that the average incremental accuracy for all the other hybrids (\textit{ResNet-18}, \textit{VGG-16}, \textit{Single-Centroid-Pred} and \textit{Remove-Centroids}) is also higher than the state-of-the-art methods.

\begin{table}
\centering
\small
\begin{tabular}{ |p{3cm}|p{1.9cm}|}
     \hline
    \textbf{Methods} & \textbf{Accuracy (\%)} \\
    \hline
    VGG-16 & 68.5\\
     \hline
    Trad-Agg & 59.2\\
    \hline
    K-means & 60.0\\
    \hline
    Single-Centroid-Pred & 68.7\\
    \hline
    Remove-Centroids & 69.0 \\
    \hline
    NCM & 58.25 \\
    \hline
    \textbf{CBCL} & \textbf{69.85} \\
 \hline
 \end{tabular}
 \caption{Effect on \textit{average incremental accuracy} by switching off each component separately in CBCL. All of the hybrids show lower performance than CBCL demonstrating each of their contribution to get the best results using CBCL}
 \label{tab:Ablation}
 \end{table}

\section{Conclusion}
\label{sec:Conclusion}

In this paper we have proposed a novel cognitively-inspired approach (CBCL) for class-incremental learning which does not store previous class data. The centroid-based representation of different classes not only produces the state-of-the-art results but also opens up novel avenues of future research, like few-shot incremental learning. Although CBCL offers superior accuracy to other incremental learners, its accuracy is still lower than single batch learning on the entire training set. Future versions of CBCL will seek to match the accuracy of single batch learning. Although, for FSIL CBCL beats the batch learning baseline.

CBCL contributes methods that may one day allow for real-world incremental learning from a human to an artificial system. Few-shot incremental learning, in particular, holds promise as a method by which humans could conceivably teach robots about important task-related objects. Our upcoming work will focus on this problem. 



\section*{Acknowledgments}
\noindent This work was supported by Air Force Office of Scientific Research contract FA9550-17-1-0017.

{\small
\bibliographystyle{ieee_fullname}
\bibliography{main}
}

\newpage
\section{Supplementary Material}
\beginsupplement
\subsection{CBCL Algorithms}
The algorithms below describe portions of the complete CBCL algorithm. Algorithm 1 is for \textit{Agg-Var} clustering (Section 3.1 in paper), Algorithm 2 is for the weighted voting scheme (Section 3.2 in paper) and Algorithm 3 is for centroid reduction technique (Section 3.3 in paper).

\begin{algorithm}[H]
\caption{CBCL: \textit{Agg-Var} Clustering}
\begin{flushleft}
        \textbf{Input:} $X=\{X^1,...,X^t\}$\Comment{feature vector sets of the training images belonging to $t$ classes}\\
        \textbf{require:} $D$\Comment{distance threshold}\\
        \textbf{Output:} $C = \{C^1,...,C^t\}$\Comment{collection of class centroid sets for $t$ classes}\\
\end{flushleft}
\begin{algorithmic}[1]
\For {$j=1$; $j \leq t$} 
\State $C^j\leftarrow\{x_1^j\}$\Comment{initialize centroids for each class}
\EndFor
\For {$j=1$; $j\leq t$}
\For {$i=2$; $i \leq N_j$}
\State $d_{min} \leftarrow \min_{l=1,..,size(C^j)} dist(c^j_{l},x^j_{i})$\Comment{distance to closest centroid}
\State $i_{min} \leftarrow \argmin_{l=1,..,size(C^j)} dist(c^j_{l},x^j_{i})$\Comment{index of the closest centroid}
\State \textbf{Set} $w^j_{i_{min}}$ to be the number of images clustered 
\State in the $i_{min}$th centroid pair of class $j$
\If {$d_{min}<D$}
\State $c^j_{i_{min}} \leftarrow \frac{w^j_{i_{min}} \times c^j_{i_{min}} + x^j_i}{w^j_{i_{min}}+1}$\Comment{update the closest centroid}
\Else
\State $C^j.append(x^j_i)$\Comment{add a new centroid for class $j$}
\EndIf
\EndFor
\EndFor
\end{algorithmic}
\end{algorithm}

\begin{algorithm}[H]
\caption{CBCL: Weighed voting scheme for classification}
\begin{flushleft}
        \textbf{Input:} x\Comment{feature vector of the test image}\\
        \textbf{require:} $n$\Comment{number of closest centroids for prediction}\\
        \textbf{require:} $C = \{C^1,...,C^t\}$\Comment{class centroids sets}\\
        \textbf{require:} $\{N_1,N_2,...,N_t\}$\Comment{Number of training images per class}\\
        \textbf{Output:} $y^*$\Comment{predicted label}
\end{flushleft}
\begin{algorithmic}[1]
\State $C^* = \{c_1,c_2,...,c_n\}$\Comment{$n$ closest centroids from set $C$}
\For{$y=1$;$y\leq t$}
\State $Pred(y) = \frac{1}{N_{y}}\sum_{j=1}^{n} \frac{1}{dist(x,c_j)}[y_j=y]$
\EndFor
\State $y^* = \argmax_{y=1,...,t} Pred(y)$
\end{algorithmic}
\end{algorithm}

\begin{algorithm}
\caption{CBCL: Centroid Reduction}
\begin{flushleft}
        \textbf{Input:}
        $C = \{C^1,...,C^t\}$\Comment{current class centroids sets}\\
        \textbf{require:} $K$\Comment{maximum number of centroids}\\
        \textbf{require:} $K_{new}$\Comment{number of centroids for new classes}\\
        \textbf{Output:} $C_{new} = \{C^{1}_{new},...,C^{t}_{new}\}$\Comment{reduced class centroids sets}\\
\end{flushleft}
\begin{algorithmic}[1]
\State $K_r = K_t+K_{new}-K$
\For {$y=1$;$y\leq t$}
\State $N^*_{y}(new) = N_y^*(1-\frac{K_r}{K_t})$
\State $C^{y}_{new} =\kmeans(n\_clusters=N^*_y(new),C^y)$ 
\EndFor
\end{algorithmic}
\end{algorithm}

\subsection{Comparison of CBCL with FearNet on CIFAR-100 Dataset}
In this section we compare CBCL against FearNet \cite{kemker18} which is another brain-inspired model for incremental learning. FearNet uses a ResNet-50 pre-trained on ImageNet for feature extraction and uses brain-inspired dual-memory model. FearNet stores the feature vectors and covariance matrices for old class images and also uses a generative model for data augmentation. For this comparison we use the evaluation metrics provided in~\cite{kemker18}. We test the model's ability to retain base-knowledge given as $\Omega_{base}=\frac{1}{T-1}\sum_{t=2}^{T}\frac{\alpha_{base,t}}{\alpha_{offline}}$, where $\alpha_{base,t}$ is the accuracy of the model on the classes learned in the first increment, $\alpha_{offline}$ is the accuracy of a multi-layer perceptron trained offline (69.9\% reported in \cite{kemker18}) and $T$ is the total number of increments. The model's ability to recall new information is evaluated as  $\Omega_{new}=\frac{1}{T-1}\sum_{t=2}^{T}\alpha_{new,t}$, where $\alpha_{new,t}$ is the accuracy of the model on the classes learned in increment $t$. Lastly, we evaluate the model on all test data as $\Omega_{all}=\frac{1}{T-1}\sum_{t=2}^{T}\frac{\alpha_{all,t}}{\alpha_{offline}}$, where $\alpha_{all,t}$ is the accuracy of the model on all the classes learned up till increment $t$. For a fair comparison, we use the ResNet-50 pre-trained on ImageNet as a feature extractor.

\begin{table}[h]
\centering
\small
\begin{tabular}{ |p{1.5cm}|p{1.1cm}|p{1.1cm}|p{1.2cm}|p{1.2cm}| }
     \hline
    \textbf{Evaluation Metric} & \textbf{FearNet} & \textbf{CBCL} & \textbf{CBCL 5-Shot} &\textbf{CBCL 10-Shot} \\
     \hline
    $\Omega_{base}$ & 0.927 & \textbf{1.025} & 0.754 & 0.830\\
     \hline
    $\Omega_{new}$ & 0.824 & \textbf{1.020} & 0.778 & \textbf{0.870}\\
     \hline
    $\Omega_{all}$ & 0.947 & \textbf{1.025} & 0.778 & 0.870 \\
 \hline
 \end{tabular}
 \bigskip
 \caption{Comparison with FearNet on CIFAR-100. $\Omega_{base}$, $\Omega_{new}$ and $\Omega_{all}$ are all normalized by the offline multi-layer preceptron (MLP) baseline (69.9\%) reported in \cite{kemker18}. A value greater than 1 means that the \textit{average incremental accuracy} of the model is higher than the offline MLP.}
 \label{tab:FearNet}
 \end{table}

\begin{figure}[t]
\centering
\includegraphics[scale=0.32]{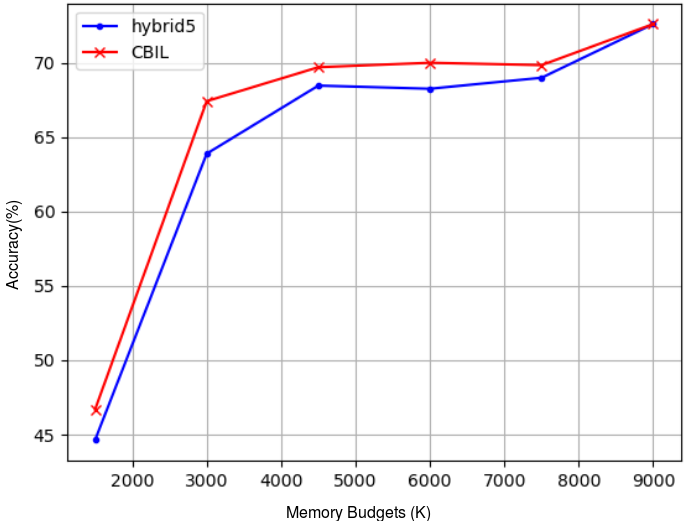}
\caption{\small Average incremental accuracy of CBCL and \textit{hybrid5} for different memory budgets (K). The difference between CBCL and \textit{hybrid5} is more prominent for smaller memory budgets.}
\label{fig:memory_budgets}
\end{figure}

Table \ref{tab:FearNet} compares CBCL with FearNet on CIFAR-100 dataset using the metrics proposed in \cite{kemker18}. We report results of CBCL on the most difficult increment setting (2 base classes and then 1 class per increment for 98 classes) for this experiment. CBCL clearly outperforms FearNet on all three metrics ($\Omega_{base}$, $\Omega_{new}$, $\Omega_{all}$) by a significant margin when using all training examples per class. For 10-shot incremental learning, CBCL outperforms FearNet (which uses all the training examples per class) on $\Omega_{new}$ but for $\Omega_{base}$ and $\Omega_{all}$ it is slightly inferior. For 5-shot incremental learning setting, the results of CBCL are inferior to FearNet (which uses all the training examples) but the change in accuracy is not drastic. It should be noted that even for 10-shot and 5-shot incremental learning settings, the MLP baseline, used during the calculation of $\Omega_{base}$, $\Omega_{new}$ and $\Omega_{all}$, has been trained on all the training data of each class in a single batch. 

We also trained a ResNet-50 for 5-shot and 10-shot learning with all the class training data available in one batch and the test accuracies for 5-shot and 10-shot learning were 8.49\% and 12.21\%, respectively. CBCL outperforms this baseline by a remarkable margin for both 5-shot and 10-shot settings, demonstrating that it is extremely effective for few-shot incremental learning setting.

\subsection{Analysis of Different Memory Budgets}
We perform a set of experiments on CIFAR-100 dataset to analyze the effect of different memory budgets on the performance of CBCL. We performed these experiments on \textit{hybrid5} as well to show the contribution of our proposed centroid reduction technique towards CBCL's performance. Figure \ref{fig:memory_budgets} compares the average incremental accuracy of CBCL and \textit{hybrid5} for different memory budgets. As expected, both CBCL and \textit{hybrid5} achieve higher accuracy for when provided higher memory budgets. Furthermore, CBCL constantly outperforms \textit{hybrid5} for all different memory budgets (except for K=9000 when there is no need for any reduction) and the performance gap increases for smaller memory budgets. This clearly shows the effectiveness of our proposed centroid reduction technique over simple removal of centroids. Furthermore, it should be noted that even for only $K=3000$ centroids CBCL's average incremental accuracy (\textbf{67.5\%}) is higher than that of the state-of-the-art methods (\cite{Wu_2019_CVPR}: 64.84\%).

\subsection{Confusion Matrices}
\begin{figure}[t]
\centering
\includegraphics[scale=0.31]{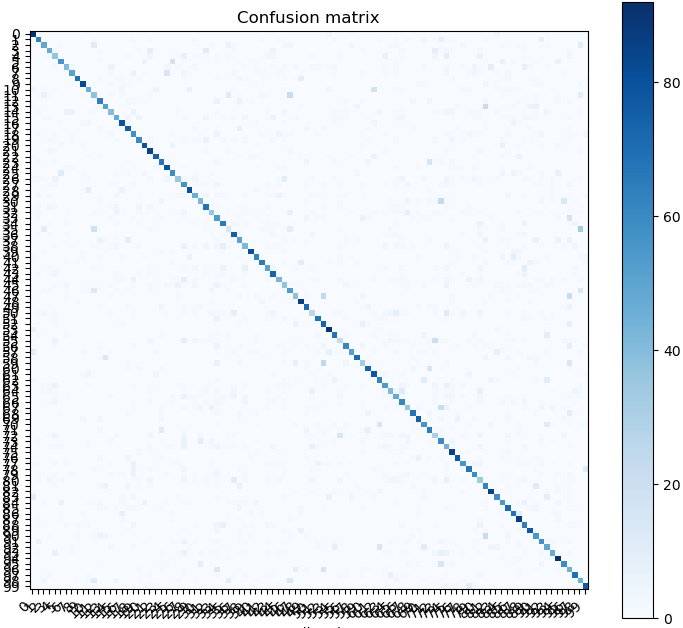}
\caption{\small Confusion matrix of CBCL on CIFAR-100 dataset with 10 classes per increment and total centroids limit of $K=$7500. The vertical axis depicts the ground truth and the horizontal axis shows the predicted labels (0-99).}
\label{fig:confusion}
\end{figure}

We further provide insight into the behavior of CBCL through the confusion matrix. Figure \ref{fig:confusion} shows the confusion matrix of CBCL on CIFAR-100 dataset when learning with 10 classes per increment with a memory budget of $K=$7500. The pattern is quite obvious that the confusion matrix of CBCL looks homogenous in terms of diagonal and off-diagonal entries depicting that CBCL does not get biased towards new or old classes and it does not suffer from \textit{catastrophic forgetting}.

\section{Hyperparameters}
CBCL only has two hyperparameters: distance threshold ($D$) and number of centroids used for classification ($n$). For all three datasets (CIFAR-100, Catltech-101 and CUBS-200-2011), $D$ was tuned to one of the values in the set $\{70,75,80,85,90\}$, although in most of the increments it was tuned to $70$ for both incremental learning and FSIL experiments. $n$ was tuned to one of the values in the set $\{1,2,...,10\}$ for incremental learning experiments but for FSIL experiments it was mostly tuned to $1$.

\subsection{Results on Caltech-101 Using Bag of Visual Words}
To show the effect of feature extractor choice on CBCL's performance, we report results on Caltech-101 dataset using bag of visual words (with SURF features \cite{Bay08}). Bag of visual words (BoVW) features are significantly inferior to CNN features on image classification tasks. Table~\ref{tab:Caltech_bag} compares CBCL using BoVW against LWM and \textit{finetuning} (FT) with 10 classes per increment. CBCL's accuracy is significantly lower than LWM and FT for the first increment (because of inferior features) and for all the other 9 increments it is either higher or slightly inferior to LWM. This shows that CBCL yields near state-of-the-art accuracy even when using inferior features. Furthermore, it should be noted that the decrease in accuracy of CBCL is still only \textbf{37.61\%} after 10 increments while for LWM and FT the decrease in accuracies are 69.52\% and and 49.36\%. These results clearly show the effectiveness of CBCL to avoid \textit{catastrophic forgetting}. 

\begin{table}[H]
\centering
\small
\begin{tabular}{ |p{1.4cm}|p{1cm}|p{1cm}|p{2.3cm}| }
     \hline
    \textbf{\# Classes} & \textbf{FT} & \textbf{LWM} & \textbf{CBCL BoVW} \\
     \hline
    10 (base) & $97.78$ & $97.78$ & $85.14\pm1.12$\\
     \hline
    20 & $59.55$ & $75.34$ & $\boldsymbol{77.84\pm1.86}$\\
    \hline
    30 & $52.65$ & $71.78$ & $69.65\pm2.21$\\
    \hline
    40 & $44.51$ & $67.49$ & $63.89\pm1.40$\\
    \hline
    50 & $35.52$ & $59.79$ & $\boldsymbol{60.30\pm1.73}$\\
    \hline
    60 & $31.18$ & $56.62$ & $\boldsymbol{57.20\pm1.37}$\\
    \hline
    70 & $32.99$ & $54.62$ & $\boldsymbol{55.25\pm0.99}$\\
    \hline
    80 & $27.45$ & $48.71$ & $\boldsymbol{51.17\pm0.84}$\\
    \hline
    90 & $28.55$ & $46.21$ & $\boldsymbol{48.13\pm0.88}$\\
    \hline
    100 & $28.26$ & $48.42$ & $47.53\pm0.69$\\
 \hline
 \end{tabular}
 \bigskip
 \caption{Comparison with FT and LWM \cite{Dhar_2019_CVPR} on Caltech-101 dataset in terms of classification accuracy (\%) with 10 classes per increment. Average and standard deviation of classification accuracies per increment are reported}
 \label{tab:Caltech_bag}
 \end{table}

\end{document}